\documentclass[runningheads]{llncs}
\usepackage{amssymb}
\usepackage{nccmath, mathtools}
\usepackage{graphicx}
\usepackage{subfig}
\usepackage{bm}
\newcommand{\etal}{\&\textit{al. }}
\begin{document}
\title{Pruning-based Topology Refinement of 3D Mesh using a 2D Alpha Mask}
%

\author{Gaétan Landreau\inst{1,2}\and
Mohamed Tamaazousti\inst{2}}

\institute{Meero, 75002 Paris, France \and Université Paris-Saclay, CEA-LIST, F-91120 Palaiseau, France}
\maketitle              
\begin{abstract}
Image-based 3D reconstruction has increasingly stunning results over the past few years with the latest improvements in computer vision and graphics. Geometry and topology are two fundamental concepts when dealing with 3D mesh structures. But the latest often remains a side issue in the 3D mesh-based reconstruction literature. Indeed, performing per-vertex elementary displacements over a 3D sphere mesh only impacts its geometry and leaves the topological structure unchanged and fixed. Whereas few attempts propose to update the geometry and the topology, all need to lean on costly 3D ground-truth to determine the faces/edges to prune. We present in this work a method that aims to refine the topology of any 3D mesh through a face-pruning strategy that extensively relies upon 2D alpha masks and camera pose information. Our solution leverages a differentiable renderer that renders each face as a 2D soft map. Its pixel intensity reflects the probability of being covered during the rendering process by such a face. Based on the 2D soft-masks available, our method is thus able to quickly highlight all the incorrectly rendered faces for a given viewpoint. Because our module is agnostic to the network that produces the 3D mesh, it can be easily plugged into any self-supervised image-based (either synthetic or natural) 3D reconstruction pipeline to get complex meshes with a non-spherical topology. 
\end{abstract}
\section{Introduction}
\label{sec:intro}

The image-based 3D reconstruction task aims at building a 3D representation of a given object/scene depicted in a set of images. From a very early age, humans learn to apprehend their surrounding 3-dimensional environment and thus have high cognitive abilities for mentally representing the whole 3D scene structure from a single image. Doing so for any vision algorithm is way more challenging since computers do not have such sensitive prior knowledge. Inferring 3D information from a lower dimensional 2D space is thus an arduous task in visual computing. Whereas literature has tackled image-based 3D reconstruction for decades in computer vision and graphics with robust and renowned techniques such as Structure-from-Motion \cite{SFM}, the latest learning-based approaches address the problem through the new prism of deep neural networks \cite{CMR,3DfaceRecons,PIFUHD}.

The single-image-based 3D reconstruction issue even brings the challenge one step above as input information is solely constrained to a single image. From a general perspective, the latest contributions in single-image 3D reconstruction chose to work with mesh structures rather than 3D point clouds or voxels since they offer a well-balanced trade-off between computational requirements and tiny 3D details retrieval. Meshes also embed a notion of connectivity between vertices, contrary to the point cloud representation where such valuable property is inherently missing.

The rendering operation somehow fills the gap between the 3D world and the 2D image plane by mimicking the optical image formation process. 
Whereas the procedure is well-known in graphics for decades, it has only been brought into computer vision learning-based approaches for a few years. Indeed, the rasterization stage involved in any rendering process is intrinsically non-differentiable (since it requires a face selection step), making its integration in any deep architecture intractable from a backward loss computational perspective. The latest progress has led a few years ago to single-image 3D reconstruction methods where 3D ground truth labels are no longer needed: supervisory signal directly comes from a differentiable renderer at the 2-dimensional image level. 

There are two main ways to update the topology of any mesh during 3D object reconstruction: by either pruning some edges/ faces or, on the other hand, by adopting the opposite strategy and thus adding edges or vertices at the correct location to generate new faces onto the mesh surface. Single-image 3D reconstruction methods that require 3D supervision already apply these techniques in their training pipeline \cite{TMN,Total3D,GEOMetrics}. However, most of the current state-of-the-art methods in self-supervised single-image 3D reconstruction -where 3D labels are thus no more needed- perform mesh reconstruction with a roughly similar approach. An Encoder-Decoder network iteratively learns to predict an elementary per-vertex displacement on a 3D template sphere to faithfully reconstruct, as better as possible, the mesh associated with the input images. Such a strategy only affects the geometry of the mesh and thus does not get consideration for its topology. Indeed, vertice position impacts edges length and dihedral face angles but leaves the overall topology unchanged: two faces sharing an edge at the beginning of the training still do so at the end. These topological considerations, yet fundamental when embracing 3D mesh structures, are often bypassed in the current self-supervised single image-based 3D reconstruction literature. We thus claim that the latest advances in differentiable rendering \cite{Soft,Pytorch3D} are informative enough to address this fundamental concept.

Our work thus brings topological considerations to the self-supervised image-based 3D reconstruction issue. From a general perspective, our method leverages the differentiable renderer from \cite{Pytorch3D} to catch up through an efficient and fast procedure the most likely mesh's faces to prune without accounting for costly 3D supervision, as done in \cite{TMN,Total3D,GEOMetrics}. As far as we are informed, no attempts in the current literature exist in this direction. Our work is thus in line with self-supervised image-based 3D reconstruction methods, while our topological refinement method is agnostic to the mesh reconstruction network used. 

We summarise our contribution through: 
\begin{itemize}
    \item A fast and efficient strategy to prune faces on a 3D mesh by only leveraging 2D alpha masks and camera pose. 
    \item An agnostic topological refinement module to the 3D mesh reconstruction network. 
\end{itemize}

\section{Related works}
\label{sec:related_works}

\noindent\textbf{Differentiable renderer} Since our work aims to be integrated within a deep architecture as an add-on module to perform complex 3D mesh reconstruction, we naturally focus on existing state-of-the-art differentiable renderers. Even though they perform much better than their differentiable counterparts, they can not be plugged into learning-based networks: there will be a network layer where back-propagation can no longer take place. OpenDR \cite{OpenDR} paved the way in 2014 regarding differentiable rendering. However, the such topic has only gained significant interest over the past few years in deep learning-based computer vision tasks. Compelling progress was reached in 2017 by Hiroharu Kato \etal with an approximated gradient-based strategy called NMR \cite{NMR}. But SoftRasterizer \cite{Soft} designed the first differentiable framework without gradient approximation through a probability-distance-based formulation whereas Chen \etal designed their differentiable renderer with foreground-background pixel consideration in their DIB-R \cite{DIBR} method. In addition to those renderers that are thus primarily designed to work with mesh, other types of renderers \cite{DVR,SDFDiff} also emerged a few years ago to address the rendering of implicit 3D shape surfaces.

\noindent\textbf{Single Image-based 3D Reconstruction} Initiating works \cite{3DRRNN,3Dold,3Dold2} related to learning-based single image 3D reconstruction extensively leveraged on 3D datasets \cite{ShapeNet,Pix3D} to let the generative network apprehends the 3D structure it must learn. These methods lack the physical image formation process during training since there is no need to consider it as soon as 3D labels are accessible. In this way, existing 3D loss functions are sufficient to predict feasible 3D mesh structures from a 3D sphere template. While tremendous works have leveraged over 3D labels, the current trend in single image-based 3D reconstruction instead tries to advantage differentiable renderers and thus limit the need for expensive 3D supervision. It led in the last few years to a new path of work called self-supervised image-based 3D reconstruction \cite{CMR,UMR,NIPS20,CVPR20} where 3D ground truth meshes are no more needed. Differentiable rendering allows to render the predicted 3D mesh onto a 2D image plane and gets a meaningful 2D supervision signal to train a mesh reconstruction network in an end-to-end way.

\noindent\textbf{Topology} Implicit-based methods spontaneously handle complex topology since any 3D object parameterises in a continuous 3-dimensional vector field where the notion of connectivity is absent. Generated surfaces do not suffer from resolution limitations as soon as the 3D space is continuously defined. Works relying on such formulation produce outstanding results but often require extensive use of 3D supervision \cite{PIFUHD}, even though the latest research achieved reconstructing 3D implicit surfaces without 3D labels \cite{DVR,ImplicitNo3D}.

The topological issue on explicit-based formulation are already addressed when it comes to supervise the mesh generation with 3D labels. Pix2Mesh \cite{Pixel2Mesh} leverages the capacity of Graph Neural Networks and their graph unpooling operation to add new vertices on the initial template mesh during training. With the same will to add a new vertex/face, GEOMetrics \cite{GEOMetrics} considers an explicit adaptive face splitting strategy to locally increase face density and thus ensure that the generated mesh will have enough detail around the most complex regions. The face splitting decision relies on local curvature consideration with a fixed threshold. These two methods adopt a progressive mesh growing strategy and thus start from a low-resolution template mesh to end up with a 3D mesh which is complex only in the most challenging regions to reconstruct.

On the other hand, Junyi Pan \etal \cite{TMN} paved the way to prune irrelevant faces onto 3D mesh surface. They introduced a face-pruning method through a 3D point cloud-based error-estimation network. While \cite{TMN} used a fixed scalar threshold to determine whether or not to prune a face, Total3D \cite{Total3D} proposes a refined version of such a method by performing edge pruning with an adaptative thresholding strategy set on 3D local considerations.

To the best of our knowledge,  such topological issue on 3D mesh structures is currently not addressed in the state of the art methods that extensively rely on 2D cues for training. Generated meshes are thus always isomorphic to a 3D sphere.  

\section{Method}
\label{sec:method}

We introduce our method and the associated framework in this section. We draft a complete overview of our methodology before digging into the implementation details of the module we designed. 

Regarding the notation, we denote by $I\in\mathbb{R}^{H\times W\times 4}$ the source RGB$\bm{\alpha}$ image, where $\bm{\alpha}$ therefore refers to the (ground-truth) alpha mask. We aim to refine the topology of a mesh \textbf{M}=(V,F) where V and F respectively stand for the set of vertices and faces. We assume such mesh was obtained from a genus-0 template shape by any single-image 3D mesh reconstruction network (fed with either the RGB image or its alpha mask counterpart). Finally, the camera pose $\theta$ is parametrized by an azimuth and an elevation angle, leaving the distance between the object and the camera fixed. 

\subsection{General overview}
\begin{figure*}[htp!]
\begin{center}
\includegraphics[width=12cm]{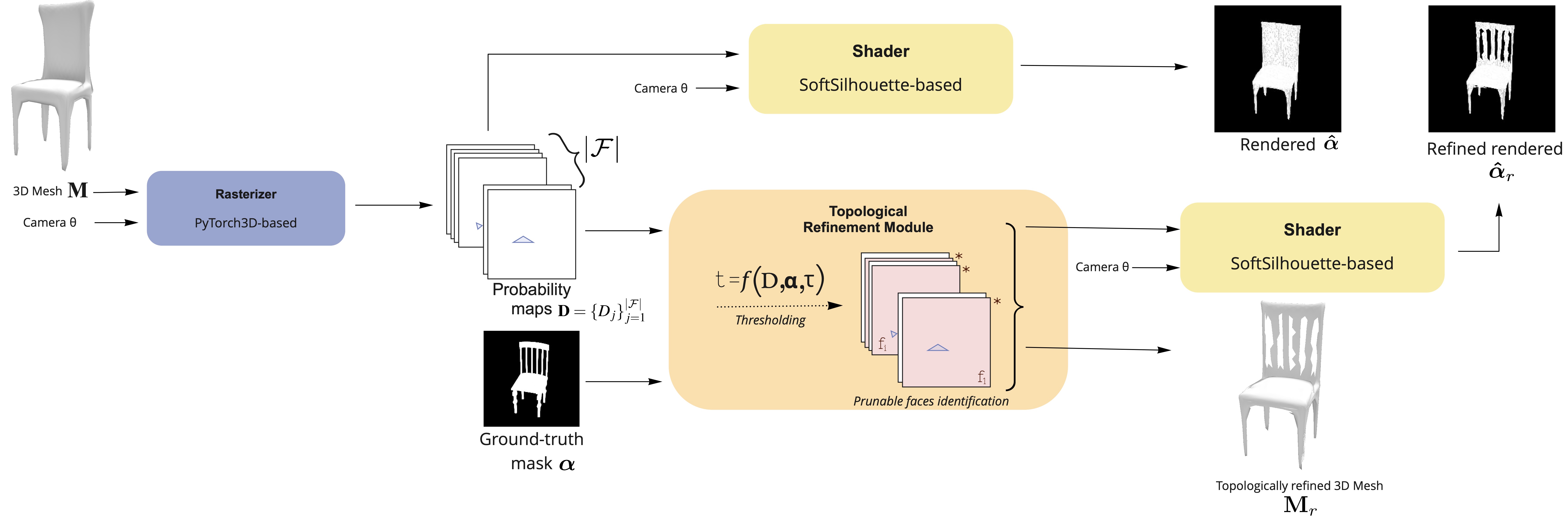}
\end{center}
    \caption{Architecture overview of our method. \textit{Based on a 3D mesh} \textbf{M} \textit{and a camera pose} $\theta$, \textit{our module leverages PyTorch3D rasterizer to detect and prune onto the mesh surface by only getting consideration for the ground-truth alpha mask $\bm{\alpha}$}.}
\label{fig:pipeline_overview}
\end{figure*}

As we extensively rely on the 2D information from $\bm{\alpha}$ (even though the 3D corresponding camera pose $\theta$ is needed) to perform topological refinement over the mesh surface, we must lean on a renderer to get back onto 2D considerations. We consider the differentiable one from PyTorch3D \cite{Pytorch3D} since it allows the generation of meaningful per-face rendered maps that one can aggregate to produce the final rendered mask. The core idea of our work is to identify the faces that were re-projected the worst onto the 2D image plane during the rasterization procedure through the prior information from $\bm{\alpha}$. Figure \ref{fig:pipeline_overview} depicts the general overview of our face-pruning method.

Detecting those faces is driven through the computation of an Intersection over Union (IoU) score between each per-face rendered map with ground-truth $\bm{\alpha}$. Those faces can then be removed from the 3D mesh surface or directly discarded in the shader stage of the renderer. Inspired by the thresholding strategy introduced in TMN \cite{TMN}, we get consideration for $t$, an adaptative threshold based on the IoU score distribution ${\gamma/\Gamma}$ and quantile $Q_{\tau}$, $\tau \in [0,1]$.

\begin{equation}
    t=Q_{\tau}({\gamma/\Gamma})
\end{equation}

In a similar fashion line to what TMN\cite{TMN} did for the thresholding strategy in their pipeline architecture, the setting of $\tau$ influences the number of pruned faces: the lower $\tau$ is, the lower the number of faces detected as wrongly projected will be.

\subsection{Implementation details.}
We implement our topological refinement strategy onto the renderer from the PyTorch3D \cite{Pytorch3D} library. The renderer's modularity offered by \cite{Pytorch3D} is worth mentioning since the entire rendering procedure can be adjusted as desired. We paid attention to the rasterization stage for its connivance with the one from SoftRasterizer \cite{Soft}. 

One of the core differences between those two frameworks in the silhouette rasterization process concerns the number of faces involved: while PyTorch3D only considers for each pixel location $p_i$ the top-\textit{K} closest faces from the camera center, SoftRasterizer equally considers all the faces of \textbf{M}.
We denote by $\mathbf{P}\in \mathbb{R}^{K\times(H\times W)}$ the intermediate probability map produced by \cite{Pytorch3D} which is highly related to the one originally introduced in \cite{Soft}. Considering any 2D pixel location $p_{i}=(x_{i};y_{i}) \in \{0,..H-1\}\times\in \{0,..W-1\} $ and the $k^{th}$ closest face $f_{k}^{i}$, the distance based probability tensor $\mathbf{P}$ is expressed through:

\begin{equation}
    \mathbf{P}[k,p_{i}]=\left(1+e^{-d(f_{k}^{i},p_{i})/\sigma}\right)^{-1} 
    \label{eq:soft_value}
\end{equation}

where $d(f_{k}^{i},p_{i})$ stands for the Euclidean distance between $p_i$ and $f_{k}^{i}$, while $\sigma$ is a hyperparameter to control the sharpness of the rendered silhouette image. Both $d$ and $\sigma$ are defined in SoftRasterizer \cite{Soft}. 

It is worth emphasizing the indexing notation of $\mathbf{P}$. Indeed, face indexes $f_{k}^{i}$ and $f_{k'}^{i'}$, $\{i,k\} \neq \{i',k'\}$ might refer to the same physical face on \textbf{M} because a rendered one is likely to cover an area larger than a single pixel. One could already build up an aggregation function to render a final predicted alpha mask from \textbf{P} but the computational cost would not be optimal. \newline

We thus introduced $\mathcal{F}$ as the set of unique faces from $\mathbf{P}$ involved in the rendering of \textbf{M}. The larger K is, the more likely the cardinality of $\mathcal{F}$ will get close to the total number of faces in the original mesh $|F|$. 

We denote by $\mathbf{D}=\{D_{j}\}_{j=1}^{|\mathcal{F}|}\in \mathbb{R}^{|\mathcal{F}|\times(H\times W)}$ the probability map tensor, as defined in \cite{Soft}, that accounts (contrary to \textbf{P}) on all the unique faces (indexed $f_{j}$) involved in the rendering process. Following Equation \ref{eq:soft_value} formulation, we have for any pixel location $p_{i}$: 

\begin{equation}
    D_{j}[p_{i}]=\left(1+e^{-d(f_{j},p_{i})/\sigma}\right)^{-1} 
    \label{eq:soft_value2}
\end{equation}
Our module status on pruning the face $f_j$ considering the degree of overlap between the ground truth $\bm{\alpha}$ and the corresponding probability map $D_{j}$. Since each face $f_{j} \in \mathcal{F}$ contributes to the final rendered, an Intersection over Union (IoU) term is computed per face: 

\begin{equation}
\begin{cases}
     \gamma_{j}=\sum_{p_{i}\in \bm{\alpha}} \min \left(  D_{j}[p_{i}] , \bm{\alpha}[p_{i}] \right) \\
     \Gamma_{j}=\sum_{p_{i}\in \bm{\alpha}} \max \left( D_{j}[p_{i}],\bm{\alpha}[p_{i}] \right)
\end{cases}
\end{equation}

The ratio $\gamma_{j}/\Gamma_{j}$ gives the well-known IoU score. We extend the computation for a single face $f_{j}$ to all the faces from $\mathcal{F}$, and denote by $\gamma/\Gamma \in \mathbb{R}^{|\mathcal{F}|}$ the complete IoU score distribution.\newline

We adopt a thresholding strategy partially inspired from \cite{TMN} and set an adaptative threshold $t$ based on statistical quantile consideration: faces with a lower IoU score than $t=Q_{\tau}(\gamma/\Gamma)$ are pruned from $\mathbf{M}$ to give a refined mesh $\mathbf{M}_{r}$.\newline

Given all these considerations, two different predictions can be made on the final rendered mask: 

\begin{equation}
\begin{cases}
     \bm{\hat{\alpha}}[p_i]=1 - \prod_{j=1}^{|\mathcal{F}|} (1 - D_{j}[p_{i}]) \\
     \bm{\hat{\alpha}}_{r}[p_i]=1 - \prod_{j=1}^{|\mathcal{F}\setminus\mathcal{F}_{p}|} (1 - D_{j}[p_{i}])
\end{cases}
\end{equation}
While $\bm{\hat{\alpha}}$ to the original predicted alpha mask (without any faces pruned), $\bm{\hat{\alpha}}_{r}$ refers to the refined predicted silhouette, with $\mathcal{F}_{p}=\{f_{p} \in \mathcal{F}| \gamma_{p}/\Gamma_{p} < t\}$
\section{Experiments}
\label{sec:experiments}

\textbf{Dataset} We extensively tested our approach on ShapeNetCore \cite{ShapeNet}. In line with the work from TMN \cite{TMN}, our experiments are thus limited to the topologically challenging "chair" class from \cite{ShapeNet}. It contains 6774 different chairs, with 1356 instances in the testing set.

\noindent\textbf{Metrics} We evaluate our method through both qualitative and quantitative considerations. We use the 2D IoU metric to assess how well the refined mesh produced by our module better matches the ground truth alpha mask compared to the topologically non-refined mesh. We also use 3D metrics with the Chamfer Distance (CD), F-Score and METRO distance to evaluate our method. The METRO criterion was introduced in \cite{metro} and reconsidered in Thibault Groueix \etal 's AtlasNet \cite{AtlasNet} work. Its use is motivated by its consideration for mesh connectivity contrary to the CD or F-score metric that only reason onto 3D point clouds distribution.

\noindent\textbf{3D Mesh generation network} Our refinement module can be integrated into any image-based 3D reconstruction pipeline and is thus agnostic to the network responsible for producing the 3D mesh. We chose to work with the meshes generated by \cite{TMN}. Since we only want to focus on face-pruning considerations, we only retrain the ResNet18 encoder and the first stage of their 3D mesh reconstruction architecture, referred to as \textit{SubNet-1} in \cite{TMN} and abbreviated as TMN in this section. The TMN architecture thus consists of a deformation network and a learnt topological modification module. It is worth mentioning the TMN \cite{TMN} architecture has been trained and used for inference with the provided ground truth labels and rendered images from 3D-R2N2 \cite{3DRRNN}. We called "Baseline" the deformation network preceding the topology modification network \cite{TMN}. The genus-0 3D mesh produced by the Baseline network comes from a 3D sphere template with 2562 vertices.

\noindent\textbf{PyTorch3D Renderer} We use the PyTorch3D \cite{Pytorch3D} differentiable renderer and set K=30 and $\sigma=5.10^{-7}$ to get the alpha mask as sharp as possible. All the 2D alpha masks, size 224x224, were obtained with the PyTorch3D renderer and have been centred. Similarly to what \cite{3DRRNN,Soft,PTN} did for the rendering silhouette masks, we considered 24 views per meshes with a fixed camera distance $d_{camera}=2.732m$ and an elevation angle set to $30^{\circ}$. The azimuth angle varies by $15^{\circ}$ increment, from $0^{\circ}$ to $345^{\circ}$. All the meshes predicted by TMN \cite{TMN} were normalised in the same way as ShapeNetCore \cite{ShapeNet}. 

We both present qualitative and quantitative results of our pruning-based method through 2D and 3D evaluation considerations. We demonstrate how effective our strategy can be by only leveraging 2D alpha masks and the renderer modularity. 

\subsection{Topological refinement evaluation - Qualitative results}

\begin{figure*}[htb]
\begin{center}
\includegraphics[width=8cm]{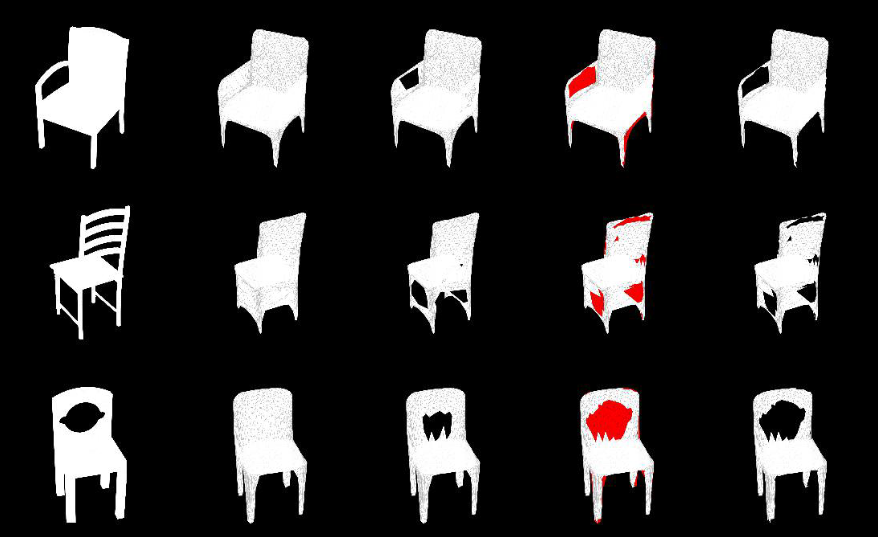}
\end{center}
    \caption{Silhouette based comparison on several instance from the ShapeNetCore test set. \textit{Faces rendered onto red regions should be pruned on 3D mesh surface} - $\tau = $ \textbf{0.05} - From left to right: Ground-Truth, Baseline, TMN \cite{TMN}, Ours with highlighted faces to prune, Ours final result.}
\label{fig:face2prune}
\end{figure*}

We first seek to highlight to what extent we can detect irrelevant faces on the 3D mesh, i.e those that might be pruned during rendering. Figure \ref{fig:face2prune} depicts the wrongly rendered faces (considered as is by our method) compared to the ground-truth alpha mask on three different chairs. Based on these 2D silhouette considerations, we achieve visually more appealing results than \cite{TMN}. 

\begin{figure*}[h!]
\begin{center}
\includegraphics[width=8.cm]{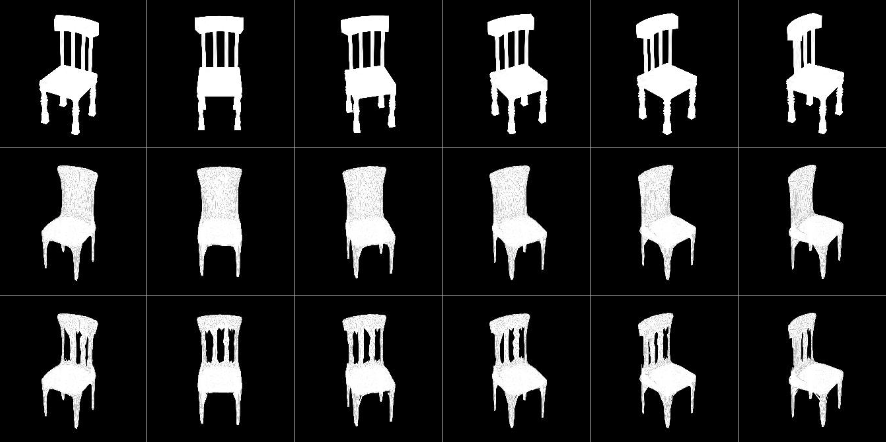}
\end{center}
    \caption{Rendered silhouette mask results on 6 viewpoints - $\tau =\textbf{0.05}$ - From top to bottom: Ground-Truth, TMN \cite{TMN},  Ours}
\label{fig:pruning_multi_view}
\end{figure*}

Figure \ref{fig:pruning_multi_view} somehow extends the later observation through 6 different viewpoints from the same chair instance. In this example, the TMN pruning module failed to detect some faces to discard. It produced the same mesh as the baseline one, while our method successfully pruned the faces that have been rendered the worst, according to the ground truth alpha mask. Pruned faces on each view are independent of the other viewpoints. 

Even the viewpoint associated with a tricky azimuth angle as the one depicted in the last column of Figure \ref{fig:pruning_multi_view} is informative enough for our module to remove the relevant faces during rendering.

\subsection{2D and 3D-based quantitative evaluation}

We compare the performances of our method through different thresholds $\tau$ in Table \ref{tab:sota_table} with the meshes produced by the Baseline network and TMN \cite{TMN}. From the 1356 inferred meshes in the ShapeNetCore \cite{ShapeNet} test set, we manually selected 50 highly challenging meshes (from a topological perspective) and rendered them through 24 different camera viewpoints with the PyTorch3D renderer. The intrinsic F-score threshold was set to 0.001. A total number of N=10.000 points have been uniformly sampled over the different meshes' surfaces to compute the 3D metrics. 

\begin{table}[htp!]
\begin{center}
\caption{2D and 3D-based metric scores comparison with the Baseline and TMN \cite{TMN} - \textit{Presented results were averaged over the 50 instance from our manually curated test set and over the 24 different viewpoints for the 3D metrics.}} 
\label{tab:sota_table}

\begin{tabular}{|l|c|c|c|c|}
\hline
\textit{Method} & 2D IoU $\uparrow$   & CD $\downarrow$  &F-Score $\uparrow$ & METRO $\downarrow$ \\ 
 \hline\hline
Baseline & 0.660  &  6.602  & 53.27  &  1.419      \\ 
\hline
TMN\cite{TMN} & 0.681  & \textbf{6.328}   & \textbf{54.23}  & \textbf{1.293} \\ 
\hline
Ours $\scriptstyle \tau=0.01$ & 0.747 & 6.541  &  53.39 &  1.418      \\
\hline
Ours $\scriptstyle \tau=0.03$ & 0.755 &6.539  & 53.39  &    1.417  \\
\hline
Ours $\scriptstyle \tau=0.05$ & 0.763 & 6.540  &   53.34 &   1.417     \\
\hline
Ours $\scriptstyle \tau=0.1$ & \textbf{0.778} & 6.551  & 53.27 &  1.416     \\
\hline
Ours $\scriptstyle \tau=0.15$ & 0.771 & 6.548  & 53.26  &    1.416   \\
\hline

\end{tabular}
\end{center}

\end{table}

Our method outperforms the learned topology modification network from TMN \cite{TMN} according to Table \ref{tab:sota_table} when compared using the 2D IoU score. It is worth re-mentioning that presented results for TMN \cite{TMN} come from the first learned topological modification network. They thus do not consider the topological refinement from the \textit{SubNet-2} and \textit{SubNet-3} networks. Whereas none of our configurations (with different $\tau$ values) overperforms TMN \cite{TMN} on 3D metrics, we stress two points: 
\begin{enumerate}
    \item Topologically refined mesh by our method always get better results than the ones produced by the Baseline. 
    \item Our face-pruning strategy only relies on a single 2D alpha mask and does not require any form of 3D-supervised compared to \cite{TMN}. 
\end{enumerate}

\begin{figure}[h!]
  \centering
  \subfloat[a][2D IoU]{\includegraphics[width=8cm]{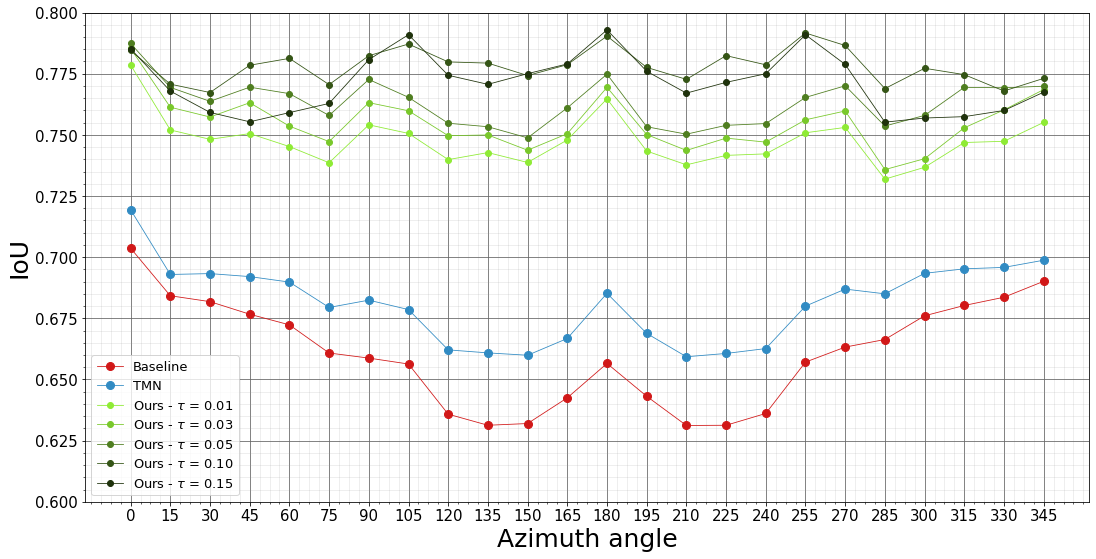} \label{fig:a}} \\
  \subfloat[b][Chamfer distance]{\includegraphics[width=8cm]{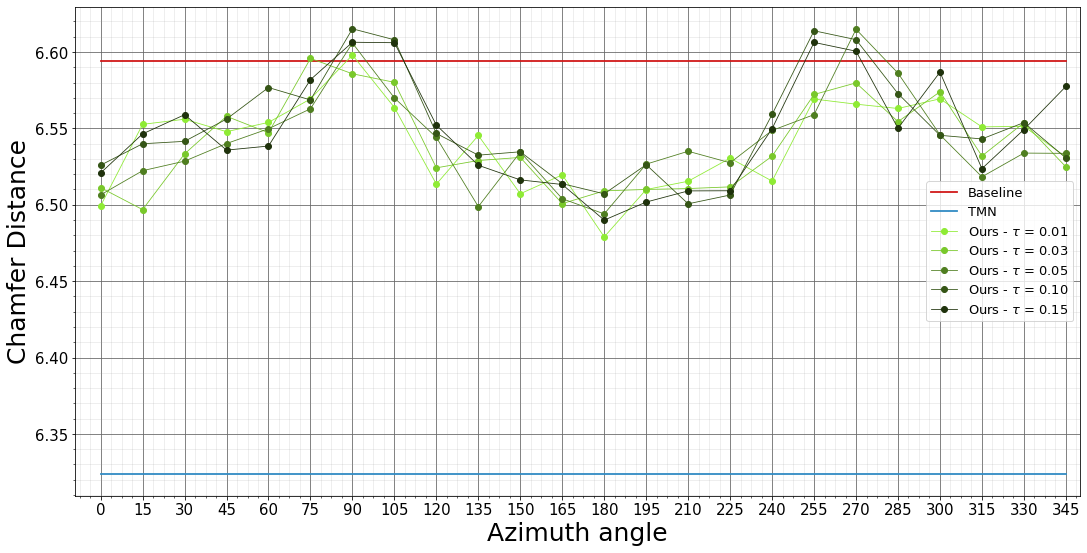} \label{fig:b}}
  \caption{Camera viewpoint influence over the 2D IoU (top, (a)) and Chamfer distance (bottom, (b) scores.} \label{fig:pruning_viewpoint_influence}
\end{figure}

Since the method we designed only relies on 2D considerations, the camera viewpoint we considered to perform the topological refinement must influence the different evaluation metrics. We show in Figure \ref{fig:pruning_viewpoint_influence} to which extent the camera pose affects both the 2D IoU and the CD scores.


Azimuth angles around the symmetrical pair $\{90^{\circ}, 270^{\circ}\}$ are more challenging since there are not as informative as the viewpoints close to $180^{\circ}$. Indeed, our method struggles to get better results than the Baseline in these cases. Our test set is imbalanced because it only contains more instances with topologically complex back parts to refine than with armrests. Our method thus slightly performs worse than the Baseline around both $90^{\circ}$ and $270^{\circ}$ angles as chairs' back complex structures are invisible from these viewpoints.   

Finally, we also quantitatively confirm the intuited impact of $\tau$ during the rendering process on the 2D IoU score: the higher $\tau$ is, the larger the number of faces we discarded.

\section{Limitations and further work}

Our method shows encouraging results in 3D meshes topological refinement through 2D alpha mask considerations but has few remaining limitations. Firstly regarding the thresholding approach we used to prune whether or not a face on the 3D mesh surface. While we require to set a fixed hyperparameter - $\tau$ - in our method as \cite{TMN} did, we align on \cite{Total3D} claims and emphasise the absolute need to rely on local 2D and 3D prior information to propose a clever and more robust thresholding strategy. Moreover, our module might also incorrectly behave on the rendered faces close to the silhouette boundary edges. 

From a broader work perspective, our method currently relies on alpha masks and thus leaves behind texture information from RGB images. While impressive 3D textured results exist with UV mapping on self-supervised image-based 3D reconstruction methods with genus-0 meshes \cite{UMR,NIPS20}, no attempts have been made to the best of our knowledge to go beyond such 0 order. Finally, since our work is agnostic to the 3D mesh reconstruction network, a natural next move would be the design of a complete self-supervised 3D reconstruction pipeline with our topological refinement module integrated. 

\section{Conclusion}
\label{sec:conclusion}
We proposed a new way to perform topological refinement onto a 3D mesh surface by only getting consideration for a 2D alpha mask. PyTorch3D \cite{Pytorch3D} rasterization framework allows our method to spot faces to discard from the mesh at almost no cost. To the best of our knowledge, no attempt exist in our line of work since both TMN \cite{TMN} and Total3D \cite{Total3D} respectively perform faces and edges pruning through 3D-supervised  neural networks. In that way, our work introduced a new research path to address the 3D mesh topology refinement issue. The agnostic design of our method allows any self-supervised image-based 3D reconstruction pipeline - based on the PyTorch3D renderer framework - to leverage the work we presented in this paper to reconstruct topologically complex meshes. We obtained consistent and competitive results from a topological perspective compared to the 3D-based pruning strategy from \cite{TMN}. 

%
%
%
\bibliographystyle{splncs04}
%

\bibliography{egbib}

\end{document}